\documentclass[10pt,conference]{IEEEtran}
\IEEEoverridecommandlockouts
\usepackage{cite}
\usepackage{amsmath,amssymb,amsfonts}
\usepackage{algorithmic}
\usepackage{graphicx}
\usepackage{textcomp}
\usepackage{xcolor}
\usepackage{booktabs}
\usepackage{url}

\def\BibTeX{{\rm B\kern-.05em{\sc i\kern-.025em b}\kern-.08em
    T\kern-.1667em\lower.7ex\hbox{E}\kern-.125emX}}
\begin{document}

\title{CuraLight: Debate-Guided Data Curation for LLM-Centered Traffic Signal Control
}

\author{
  \IEEEauthorblockN{
    Qing Guo\IEEEauthorrefmark{1},
    Xinhang Li\IEEEauthorrefmark{1},
    Junyu Chen\IEEEauthorrefmark{1},
    Zheng Guo\IEEEauthorrefmark{1},\\
    Shengzhe Xu\IEEEauthorrefmark{1},
    Lin Zhang\IEEEauthorrefmark{1}\textsuperscript{,}\IEEEauthorrefmark{2},
    Lei Li\IEEEauthorrefmark{1}
  }
  \IEEEauthorblockA{\IEEEauthorrefmark{1} School of Artificial Intelligence, Beijing University of Posts and Telecommunications, Beijing, China}
  \IEEEauthorblockA{\IEEEauthorrefmark{2} Beijing Big Data Center, Beijing, China}
  % \IEEEauthorblockA{\IEEEauthorrefmark{3} Eastern Institute of Technology, Ningbo, China}
  % \IEEEauthorblockA{
  %   Emails: \{youshen,lixinhang,junyuchen,gzheng,zhanglin,leili\}@bupt.edu.cn;\
  %   li\_xiaocong@simtech.a-star.edu.sg;\ 
  %   eleaam@nus.edu.sg
  % }
  \thanks{Corresponding author: Lei Li (leili@bupt.edu.cn).}
}
% \and
% \and
% \and
% \and
% \and

\maketitle

\begin{abstract}
As a core component of intelligent transportation systems (ITS), traffic signal control (TSC) mitigates congestion, reduces emissions, and improves travel-time reliability. Recent TSC methods based on rules, reinforcement learning (RL), and large language models (LLMs) have improved adaptivity across urban networks. Yet key limitations persist: rule-based methods and RL methods provide limited interpretability of timing actions; LLM-based methods face challenges in selecting high-dimensional timings at heterogeneous intersections; and because LLMs are typically trained with far less task-aligned interactive data than RL, intersection-specific operational knowledge remains limited. This paper presents CuraLight, an LLM-centered framework in which an RL agent assists the fine-tuning of an LLM agent. RL exploration gathers traffic states and actions, enhanced prompting constructs imitation pairs for fine-tuning the base model, timing decisions are structured to elicit interpretable preferences, and a multi-LLM ensemble deliberation system conducts adversarial debates that defend each RL-filtered phase action and consolidates a consensus outcome for RL-assisted fine-tuning. Evaluation in SUMO across heterogeneous networks from Jinan (17 intersections), Hangzhou (19), and Yizhuang (177) shows improvements over recent state of the art: ATT by about 5.34\%, AQL by about 5.14\%, and AWT by about 7.02\%. Code and configurations are available at \url{https://anonymous.4open.science/r/CuralightCode-6437/}.
\end{abstract}

\begin{IEEEkeywords}
traffic signal control, imitation fine-tuning, multi-agent debate, heterogeneous intersections
\end{IEEEkeywords}

\section{Introduction}
Recent advances in AI and networked sensing are accelerating Intelligent Transportation Systems, expanding the capacity to observe traffic states and coordinate operations at scale. Traffic signal control is central to allocating intersection capacity and improving travel-time reliability. However, many methods still focus on standard intersections or fixed-time settings, and dynamic timing is often studied at limited scales. Compared with reinforcement learning, LLM-based TSC methods typically rely on far fewer environment interactions, leading to insufficient scene-specific learning and brittle decisions under demand shifts and mild heterogeneity. Addressing these issues is crucial for scalable, adaptive, and interpretable urban TSC.

Reinforcement learning (RL) has been widely applied to traffic signal control (TSC). Jiang et al. proposed UniComm for multi-intersection coordination via universal communication\cite{ijcai2022p535}. Gu et al. introduced $\pi$-Light, representing policies as programmatic rules for interpretable control and improved transfer\cite{Gu_Zhang_Liu_Gao_Li_Zhou_2024}. Yu et al. developed a decentralized RL framework that couples signal control with bus priority and supports model reuse across networks\cite{YU2023104281}. Despite progress, action-level interpretability and cross-scenario generalization remain fragile.

LLM-based methods leverage language-model reasoning for transparency and transfer. Da et al. proposed PromptGAT to inject LLM-derived dynamics knowledge for grounded action transformation\cite{Da_Gao_Mei_Wei_2024}. Lai et al. introduced LLMLight, using a specialized backbone to perceive textualized states and decide signal phases via prompted chain-of-thought\cite{LLMlight}. Yuan et al. presented CoLLMLight for cooperative network-wide control with spatiotemporal graph reasoning\cite{yuan2025collmlightcooperativelargelanguage}. However, evaluations often assume non-heterogeneous intersections and predefined phase sets, leaving green durations fixed and heterogeneous phasing requirements insufficiently addressed.

\begin{figure*}[t]
\centering
\includegraphics[width=\textwidth]{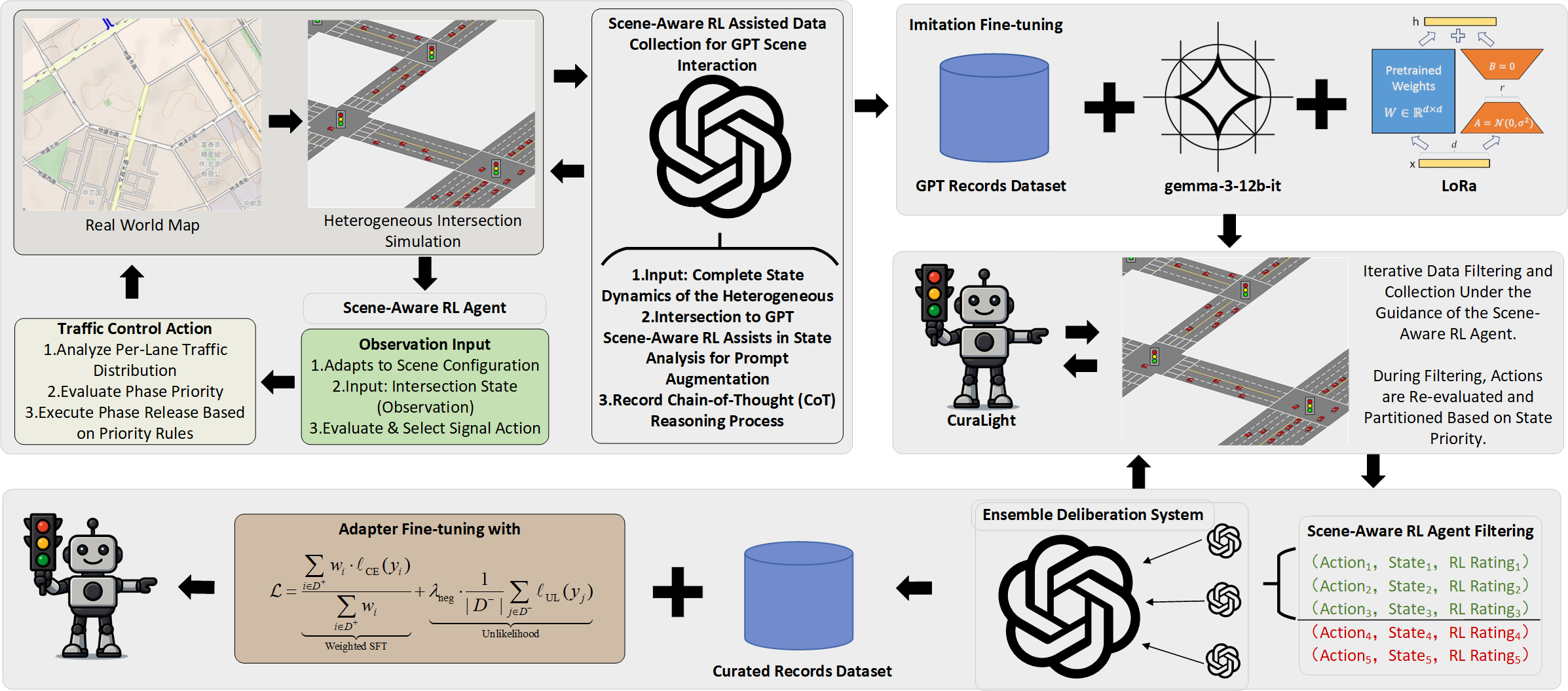}
\caption{Overview of CuraLight. RL agent assists GPT in exploring urban networks and collecting interaction trajectories for LoRA-based imitation fine-tuning of the CuraLight LLM agent. An RL-based selector and a multi-LLM ensemble deliberation system then evaluate and rank candidate signal-timing actions, providing prioritized supervision for scenario interaction data collection, filtering, and adapter training.}
\label{fig:framework}
\end{figure*}

Multi-agent deliberation and judge-ensemble methods improve reliability through cross-checking. Liang et al.\ proposed Multi-Agent Debate (MAD)\cite{liang-etal-2024-encouraging}; Estornell and Liu et al.\ studied theory-grounded debate with stabilizing interventions\cite{NEURIPS2024_32e07a11}; Jung et al.\ introduced Trust-or-Escalate with cascaded judges\cite{jung2025trust}. Existing studies mainly target single-turn tasks, whereas traffic signal control requires per-timestep selection among many phase/timing candidates under real-time constraints, motivating deliberation mechanisms tailored to sequential control.

This paper presents CuraLight, a traffic signal control agent based on a large language model trained with a diffusion-based RL assistant and a multi-LLM ensemble deliberation system (Fig.~\ref{fig:framework}). The RL assistant explores heterogeneous networks and provides reference actions and preference signals, while the deliberation system evaluates candidate timing actions and yields priority-aware supervision for adapter fine-tuning. Our contributions are:
\begin{itemize}
    \item An RL-assisted data generation pipeline is developed, where a diffusion-based RL agent explores the phase and timing space and collaborates with GPT to generate interaction trajectories, which are converted into prompt and answer pairs for LoRA-based imitation fine-tuning.
    \item A multi-LLM ensemble deliberation system is introduced to evaluate and decompose candidate signal timing actions, enabling priority-aware collection, ranking, filtering, and adapter training on scenario-specific datasets.
    \item Experiments in SUMO on heterogeneous real-world networks from Jinan (17 intersections), Hangzhou (19), and Yizhuang (177) demonstrate that CuraLight outperforms recent methods, improving ATT by about 5.34\%, AQL by about 5.14\%, and AWT by about 7.02\%, while providing interpretable LLM-driven timing decisions.
\end{itemize}

The rest of this paper is organized as follows: Section II presents heterogeneous intersection modeling and the RL-assisted training pipeline with multi-LLM ensemble deliberation for priority-aware fine-tuning. Section III reports experimental results. Section IV concludes the paper.

\section{Heterogeneous Intersection Modeling}
\begin{figure}[h]
\centering
\includegraphics[width=\columnwidth]{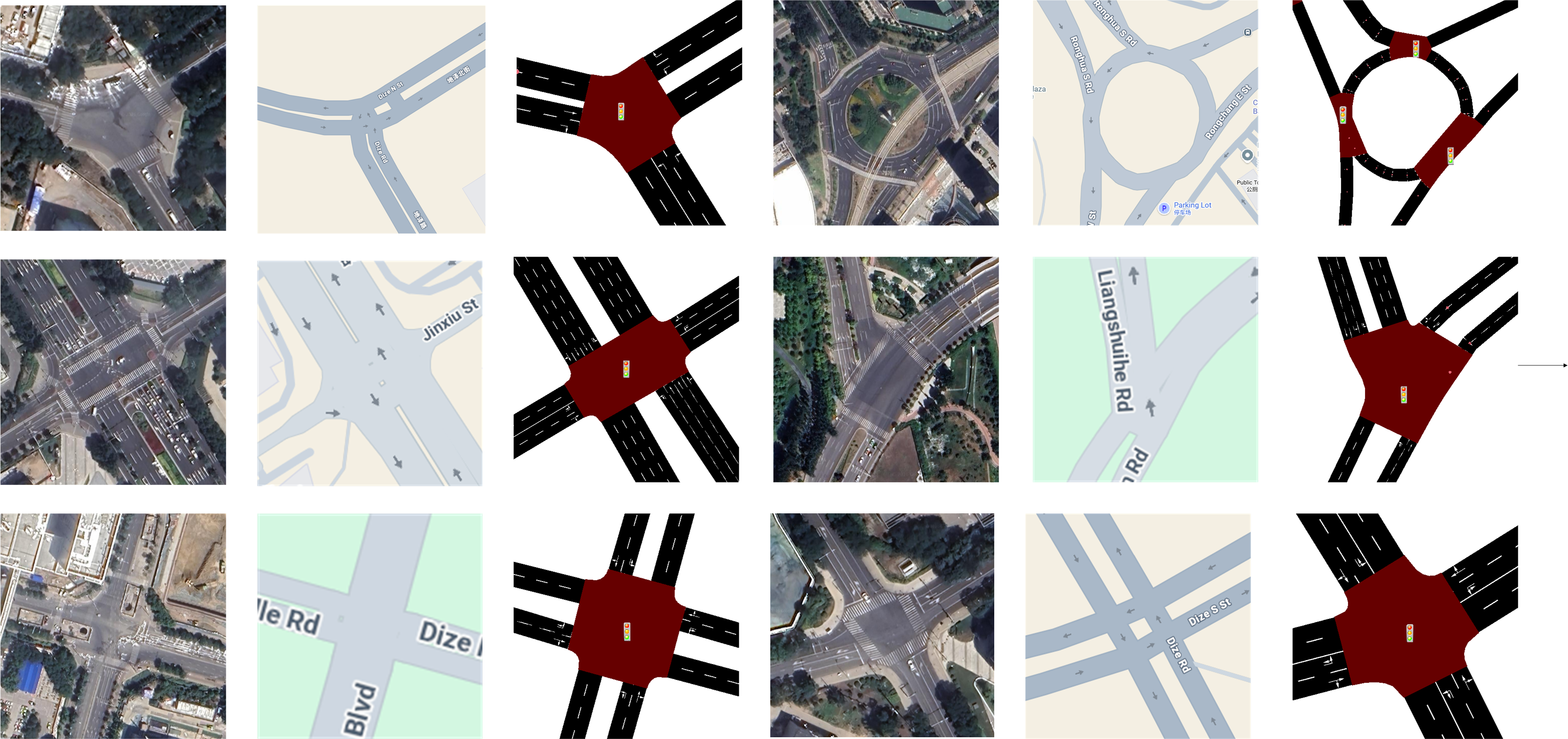}
\caption{Real-to-Simulation Modeling of Heterogeneous Intersections}
\label{fig:Intersection}
\end{figure}

As shown in Fig.~\ref{fig:Intersection}, real-world urban road networks contain intersections with diverse geometries and lane configurations, leading to heterogeneous lane-level control requirements. We adopt a unified lane-based model for subsequent state and action design.

\textit{Definition 1 (Road network).}
The urban road network is modeled as a directed graph $\mathcal{G}=(\mathcal{V},\mathcal{E})$, where $\mathcal{V}=\{1,2,\dots,I\}$ is the set of intersections and $\mathcal{E}\subseteq \mathcal{V}\times\mathcal{V}$ is the set of directed roads. Each road $(i,j)\in\mathcal{E}$ consists of a lane set $\mathcal{L}_{ij}=\{\ell_{ij}^{1},\ell_{ij}^{2},\dots\}$ with a fixed travel direction from $i$ to $j$.

\textit{Definition 2 (Traffic movements).}
At intersection $i\in\mathcal{V}$, incoming and outgoing lanes are denoted by $\mathcal{U}_i=\{u_i^{1},\dots,u_i^{n_i^{\text{in}}}\}$ and $\mathcal{O}_i=\{o_i^{1},\dots,o_i^{n_i^{\text{out}}}\}$, where $n_i^{\text{in}}=|\mathcal{U}_i|$ and $n_i^{\text{out}}=|\mathcal{O}_i|$ vary across intersections. A movement is a feasible connection $m=(u_i^{j},o_i^{k})$ with $u_i^{j}\in\mathcal{U}_i$ and $o_i^{k}\in\mathcal{O}_i$. The movement set is $\mathcal{M}_i\subseteq \mathcal{U}_i\times\mathcal{O}_i$, and $M_i=|\mathcal{M}_i|$.

\textit{Definition 3 (Signal phases and durations).}
A phase at intersection $i$ is a subset $P_i^{q}\subseteq \mathcal{M}_i$ of mutually compatible movements. The phase set is $\mathcal{P}_i=\{P_i^{1},\dots,P_i^{J_i}\}$, where $J_i$ is determined by movement conflicts. Signal control proceeds in decision steps $t=0,1,2,\dots$. At each step $t$, the controller selects a phase index $p_i(t)\in\{1,\dots,J_i\}$ and a green duration $D_i(t)$ with bounds $D_i^{\min}\le D_i(t)\le D_i^{\max}$. The pair $(p_i(t),D_i(t))$ defines the signal timing action.

\subsection{RL Agent Assisted GPT Interaction}

\subsubsection{Diffusion based RL Assistant for TSC}
Each signalized intersection $i$ is assisted by a diffusion-based RL controller. At step $t$, a short history of lane-level observations is encoded into a compact state representation $h_i(t)$. The timing action is $a_i(t)=\big(p_i(t),D_i(t)\big)$, where $p_i(t)\in\{1,\dots,J_i\}$ is a phase index selected by a \emph{Pressure-Based Selector} and $D_i(t)\in[D_i^{\min},D_i^{\max}]$ is the green duration. Conditioned on $(h_i(t),p_i(t))$, a diffusion policy $\pi_\theta\!\big(D\mid h_i(t),p_i(t)\big)$ generates duration proposals, while a \emph{History Experience Pool} retrieves a small set of previously executed durations as anchors.

\begin{figure}[h]
\centering
\includegraphics[width=\columnwidth]{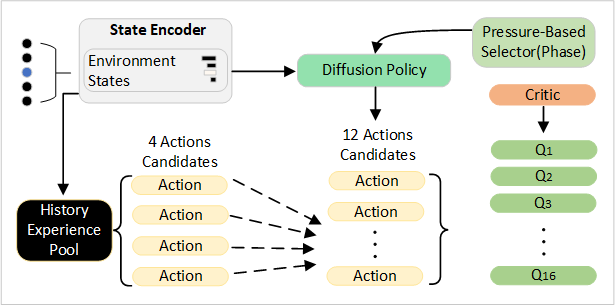}
\caption{Overview of the Diffusion-Based RL Assistant with Pressure-Based Phase Selection}
\label{fig:hybrid_diffusion_agent}
\end{figure}

During online control (Fig.~\ref{fig:hybrid_diffusion_agent}), the selector first determines $p_i(t)$ from $h_i(t)$. A candidate set of durations is then formed as
\begin{equation}
\mathcal{D}_i(t)=\mathcal{D}_{i,\mathrm{hist}}(t)\cup\mathcal{D}_{i,\mathrm{diff}}(t),
\end{equation}
where $\mathcal{D}_{i,\mathrm{hist}}(t)$ contains several history anchors and $\mathcal{D}_{i,\mathrm{diff}}(t)$ contains diffusion samples drawn from $\pi_\theta(\cdot\mid h_i(t),p_i(t))$. A distributional critic $Q_\psi\!\big(h_i(t),p_i(t),D\big)$ evaluates each candidate and selects
\begin{equation}
D_i^{\star}(t)=\arg\max_{D\in\mathcal{D}_i(t)} \;\mathbb{E}\!\left[Q_\psi\big(h_i(t),p_i(t),D\big)\right].
\end{equation}
The executed timing action is $\big(p_i(t),D_i^{\star}(t)\big)$, producing high-quality phase--duration trajectories over heterogeneous intersections.

\subsubsection{RL Assisted Guidance for LLMs}
The diffusion-based RL assistant also guides the LLM traffic signal controller. At step $t$ and intersection $i$, it provides a reference proposal $a_i^{\mathrm{RL}}(t)=\big(p_i^{\mathrm{RL}}(t),D_i^{\mathrm{RL}}(t)\big)$, where $p_i^{\mathrm{RL}}(t)$ is selected by the \emph{Pressure-Based Selector} and $D_i^{\mathrm{RL}}(t)$ is generated/selected by the diffusion duration module with the critic. The proposal is injected into the LLM prompt. RL trajectories are further summarized by (i) a predictive rollout after applying $a_i^{\mathrm{RL}}(t)$ yielding $\tilde{\mathbf{q}}_i(t+\Delta)$, and (ii) a historical time--queue mapping $D_i^{\ast}(\mathbf{q})$.

The critic provides preference scores for LLM decisions: for each $a_i^{\mathrm{LLM}}(t)$ under $h_i(t)$, it computes
\begin{equation}
Q_\psi\big(h_i(t),a_i^{\mathrm{LLM}}(t)\big),
\end{equation}
which is converted into a priority label or sampling weight. These priorities rank, filter, and subsample the LLM interaction dataset so that higher-quality, intersection-specific timing plans are retained more frequently for adapter fine-tuning.

\subsubsection{Priority Aware LoRA Imitation Fine Tuning}
Interaction data are organized as priority-scored prompt--answer pairs
\begin{equation}
\mathcal{D}=\{(x_n,y_n,s_n)\}_{n=1}^{N},
\end{equation}
where $x_n$ is the structured prompt describing the heterogeneous intersection state and RL reference signals, $y_n$ is the corresponding reasoning and phase--duration decision, and $s_n$ is a scalar priority derived from the RL critic. A filtered subset $\mathcal{D}_{\text{high}}=\{(x_n,y_n,s_n)\mid s_n\ge \kappa\}$ is used first, and lower-priority samples in $\mathcal{D}\setminus \mathcal{D}_{\text{high}}$ are introduced later via score-based sampling.

The backbone language model is the instruction-tuned Gemma-3-12B-it model. Low-rank adaptation (LoRA) injects traffic signal control knowledge by updating only adapter matrices with a standard autoregressive loss on the filtered data. The adapted LLM is referred to as the CuraLight agent.

During environment interaction, the CuraLight agent is trained under the guidance of both the diffusion-based RL assistant and a multi-LLM ensemble deliberation system. At each decision step $t$ and intersection $i$, a set of candidate phase--duration actions $\{a_i^{(k)}(t)\}_{k}$ is first evaluated by the RL assistant, and clearly inferior actions are removed by an RL-based pre-filter, leaving $\tilde{\mathcal{A}}_i(t)$.
\subsection{Training CuraLight with RL Filtering and Multi-LLM Ensemble Review}
\begin{figure}[h]
\centering
\includegraphics[width=\columnwidth]{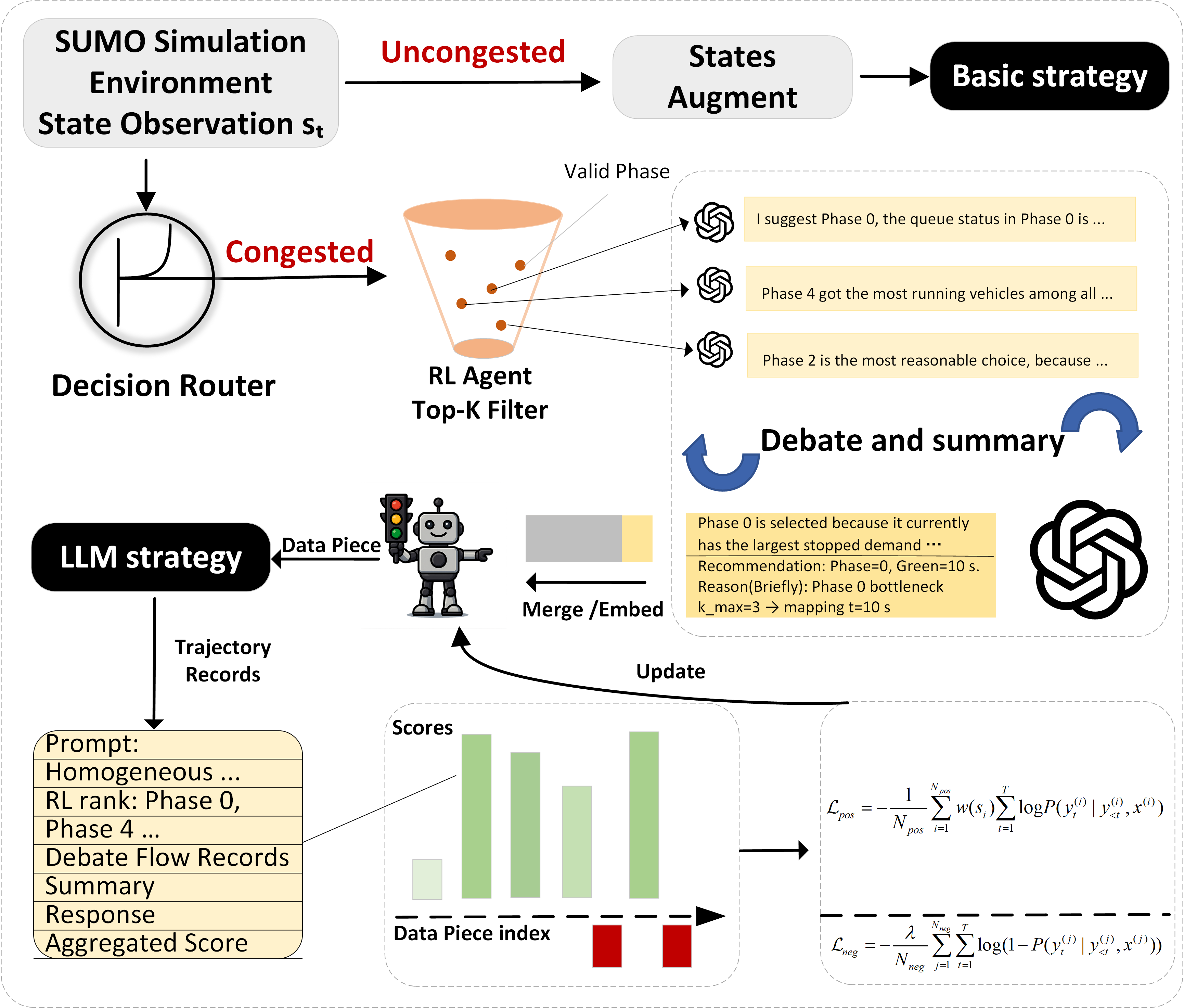}
\caption{Training Pipeline of CuraLight with RL Top-$K$ Filtering and Multi-LLM Deliberation}
\label{fig:Ensemble system}
\end{figure}

\textit{Preference scoring module.}
The consensus LLM assigns a relative preference score to each $a\in\tilde{\mathcal{A}}_i(t)$, yielding $r_i^{\mathrm{LLM}}(a)$. These scores are combined with RL critic values $q_i^{\mathrm{RL}}(a)$ to obtain a unified priority
\begin{equation}
s_i(a)=\alpha\, f\big(q_i^{\mathrm{RL}}(a)\big)+(1-\alpha)\, g\big(r_i^{\mathrm{LLM}}(a)\big),
\end{equation}
where $f(\cdot)$ and $g(\cdot)$ are normalized transformations and $\alpha\in[0,1]$ balances RL evaluation and ensemble preference. When CuraLight selects $a_i^{\mathrm{CL}}(t)$, the transition is stored with priority $s_i\big(a_i^{\mathrm{CL}}(t)\big)$. If an action has not passed through the ensemble deliberation, it is assigned a negative priority.

\textit{Second stage adapter fine-tuning.}
The unified priority scores are used as weights in a second stage of adapter fine-tuning. Trajectories are partitioned into $D^{+}$ and $D^{-}$ by the sign of their priority scores. Following \cite{mukherjee2025offline}, we optimize
\begin{equation}
\mathcal{L}
= \frac{\sum_{i \in D^+} w_i \cdot \ell_{\mathrm{CE}}(y_i)}{\sum_{i \in D^+} w_i}
+ \lambda_{\mathrm{neg}} \cdot \frac{1}{|D^-|} \sum_{j \in D^-} \ell_{\mathrm{UL}}(y_j),
\end{equation}
where $w_i\ge 0$ is obtained from the unified priority score, $\ell_{\mathrm{CE}}(\cdot)$ is the cross-entropy loss on $y_i$, and $\ell_{\mathrm{UL}}(\cdot)$ is the unlikelihood loss on $D^-$. $\lambda_{\mathrm{neg}}$ controls the repulsion strength. After convergence, the updated adapters are merged into the CuraLight agent, completing the second training stage that combines RL filtering, multi-LLM ensemble review, and priority-aware imitation for traffic signal control.

\textit{RL based pre-filtering.}
For each $a_i^{(k)}(t)$, the RL critic computes an approximate value and discards actions dominated under the current congestion pattern. The remaining set $\tilde{\mathcal{A}}_i(t)$ is forwarded to the ensemble deliberation system.

\textit{Defender LLMs.}
Each $a_i^{(k)}(t)\in\tilde{\mathcal{A}}_i(t)$ is assigned to a defender LLM, which receives the structured intersection state, competing actions, and historical summaries, and outputs a concise justification focusing on queue dissipation, balance across approaches, and anticipated network-level effects.

\textit{Consensus LLM.}
A consensus LLM aggregates defender arguments with the shared traffic context and produces a condensed comparison of candidates. This text is appended to the CuraLight prompt as a hidden auxiliary field: it is accessible to the model but cannot be directly quoted in the final explanation, encouraging internalization rather than copying.

\begin{table*}[t]
\centering
\caption{Results on \textbf{Jinan} (JN-1/2/3) and \textbf{Hangzhou} (HZ-1/2) across three metrics. All values are rounded to two decimal places. The best results are in \textbf{bold}, the second-best are \underline{underlined}, and the third-best are \underline{\underline{double underlined}}.}
\label{tab:comparison-all-rounded}
\begingroup
%\fontsize{10pt}{12pt}\selectfont 
\setlength{\tabcolsep}{4pt}      % Adjust column spacing
\renewcommand{\arraystretch}{0.94} % Adjust row spacing
\begin{tabular}{@{}l*{5}{ccc}@{}}
\toprule
Method & \multicolumn{3}{c}{JN-1} & \multicolumn{3}{c}{JN-2} & \multicolumn{3}{c}{JN-3} & \multicolumn{3}{c}{HZ-1} & \multicolumn{3}{c}{HZ-2} \\
\cmidrule(r){2-4}\cmidrule(lr){5-7}\cmidrule(lr){8-10}\cmidrule(lr){11-13}\cmidrule(l){14-16}
& ATT & AQL & AWT & ATT & AQL & AWT & ATT & AQL & AWT & ATT & AQL & AWT & ATT & AQL & AWT \\
\midrule
\multicolumn{16}{c}{\textbf{Transportation Methods}} \\
\midrule
Random           & 538.07 & 45.10 & 431.61  & 500.11 & 56.63 & 403.28  & 520.62 & 56.89 & 415.73 & 549.15 & 79.56 & 412.62  & 579.36 & 83.66 & 432.14 \\
FixedTime        & 354.94 & 27.71 & 245.06  & 363.13 & 33.13 & 252.00  & 371.43 & 36.47 & 259.37  & 473.00 & 70.32 & 311.45  & 425.75 & 66.47 & 267.11  \\
MaxPressure      & 199.26 & 14.35 & 96.68  & 214.72 & 16.96 & 107.22 & 278.03 & 33.35 & 175.47  & 257.82  & \underline{21.03} & 112.05  & 313.02 & 38.09 & 170.36  \\
\midrule
\multicolumn{16}{c}{\textbf{RL Methods}} \\
\midrule
MPLight          & 285.70 & 25.21 & 178.55  & 304.76 & 29.99 & 193.77  & 302.58 & 36.17 & 191.36  & 387.39 & 45.64  & 226.71  & 400.17 & 50.99 & 238.35 \\
AttendLight      & 319.22 & 67.05 & 246.16  & 348.91 & 69.88 & 295.60  & 446.17 & 72.81 & 383.95  & 491.44 & 113.97  & 389.20  & 411.16 & 59.45 & 309.19  \\
PressLight       & 333.63 & 27.06 & 220.03  & 335.98 & 36.79 & 225.83  & 331.65 & 38.12 & 220.51  & 378.67 & 45.46  & 222.43  & 420.83 & 63.49 & 263.62 \\
CoLight          & 532.24 & 50.49 & 426.82  & 394.93 & 47.61 & 289.64  & 449.21 & 55.60 & 346.09  & 483.14 & 73.67  & 337.13  & 458.43 & 87.70 & 311.50  \\
Efficient-CoLight& 380.94 & 36.81 & 273.61  & 403.01 & 42.44 & 288.57  & 402.07 & 44.78 & 288.11  & 446.45 & 62.61  & 293.90 & 454.47 & 67.57 & 295.01  \\
Advanced-CoLight &  362.96 & 69.22 & 296.70  & 309.55 & 73.77 & 248.83  & 380.06 & 72.16 & 309.39  & 438.32 & 85.67  & 310.35  & 429.34 & 89.46 & 284.26  \\
$\pi$-Light     &  \underline{\underline{164.64}} & 13.96 & 68.04  & 202.89 & 24.71 & 87.28  & 226.37 & 30.07 & 104.64  & 328.95 & 33.08  & 153.07  & 384.65 & 44.38 & 189.86  \\
UniTSA &  171.45 & 12.82 & 71.59  & 186.75 & \underline{15.74} & \underline{\underline{84.45}}  & \underline{\underline{200.75}} & \underline{18.18} & \underline{\underline{96.27}}  & 255.48 & 23.80  & 115.13  & \underline{\underline{277.57}} & 29.18 & \underline{\underline{134.52}}  \\
\midrule
\multicolumn{16}{c}{\textbf{Large-Scale AI Models}} \\
\midrule
Qwen3-32B      & 326.47 & 39.57 & 248.02 & \underline{\underline{185.09}} & 17.57 & 90.07 & 205.30 & 21.56 & 208.28 & 251.52 & 22.23 & 113.60 & 277.81 & 28.47 & 136.43 \\
Llama-3-70B       & 158.61 & \underline{\underline{11.49}} & 64.28  & 200.51 & 18.42 & 104.30  & 223.17 & 22.86 & 125.90 & 247.14 & 21.76 & 110.12 & 279.79 & \underline{\underline{27.92}} & 136.96 \\

Llama-4-Scout     & 156.93 & 12.38 & 61.64  & \underline{184.10} & 17.97 & 84.59  & 215.57 & 25.11 & 122.03 & \underline{\underline{244.95}} & 22.04 & \underline{\underline{106.35}} & 286.87 & 30.70 & 141.55  \\
Gemma-3-12b-it  & 163.58 & 13.49 & 70.23  & 242.15 & 20.35 & 94.45  & 273.66 & 37.21 & 150.28  & 269.42 & 30.13 & 130.64 & 315.42 & 39.78 & 165.25 \\
Gemma-3-27b-it  & 158.46 & 12.78 & 65.45  & 223.06 & \underline{\underline{16.95}} & \underline{84.41}  & \underline{197.32} & \underline{\underline{19.78}} & \underline{95.79}  & 267.44 & 24.52 & 126.67 & 300.93 & 31.36 & 152.32 \\
DeepSeek-R1-14B    & 207.25 & 18.92 & 115.23  & 198.30 & 17.68 & 100.31  & 241.17 & 23.86 & 138.10 & 274.66 & 33.01 & 139.82 & 299.13 & 31.43 & 150.38 \\
DeepSeek-R1-32B & 169.56 & 12.88 & 73.55  & 199.82 & 18.05 & 103.37  & 280.32 & 31.02 & 172.80  & 248.72 & \underline{\underline{21.69}} & 111.84 & 289.11 & 30.15 & 144.62 \\

GPT-5-mini    & \underline{149.23} & \underline{11.05} & \textbf{53.16}  & \textbf{176.52} & 17.47 & \textbf{81.75}  & 264.11 & 33.14 & 174.60 & \underline{239.34} & 24.47 & \underline{101.33} & \underline{272.58} & \underline{27.34} & \underline{127.69} \\
\midrule
\multicolumn{16}{c}{\textbf{LLM-Based Methods}} \\
\midrule
LLMLight         & 235.01 & 18.13 & 134.20 & 343.01 & 34.20 & 250.47  & 450.40 & 51.91 & 369.22  & 328.34 & 34.17  & 190.71  & 334.85 & 38.10 & 191.67  \\
Traffic-R1         & 308.88 & 33.65 & 239.46 & 323.01 & 27.92 & 263.98  & 287.78 & 35.43 & 228.91  &  343.72 & 40.52  & 222.25  & 405.43 & 55.16 & 280.87  \\
CuraLight      & \textbf{148.95} & \textbf{10.86} & \underline{54.60} & 212.19 & \textbf{13.06} & 116.60  & \textbf{182.68} & \textbf{17.81} & \textbf{85.25}  &  \textbf{238.74} & \textbf{20.65}  & \textbf{99.68}  & \textbf{262.14}& \textbf{25.03} & \textbf{118.80}  \\
\bottomrule
\end{tabular}
\endgroup
\end{table*}

\section{Experiments and Results}
\subsection{Experiments Settings}
\subsubsection{Performance metrics}ATT (s) denotes average travel time; AQL (m) denotes the network-wide average queue length; and AWT (s) denotes the average waiting time at intersections.

\subsubsection{Compared Models}
Baselines include transportation methods (Random, FixedTime~\cite{Koonce2008}, MaxPressure~\cite{VARAIYA2013177}), deep RL families (PressLight~\cite{10.1145/3292500.3330949}, MPLight~\cite{Chen_Wei_Xu_Zheng_Yang_Xiong_Xu_Li_2020}, CoLight~\cite{10.1145/3357384.3357902}, Efficient-CoLight~\cite{Wu2021EfficientPI}, Advanced-CoLight~\cite{advanced_xlight}, AttendLight~\cite{NEURIPS2020_29e48b79}, $\pi$-Light\cite{Gu_Zhang_Liu_Gao_Li_Zhou_2024}, UniTSA\cite{UniTSA} ), Large-scale AI Models (Qwen3-32B, Llama-3-70B, Llama-4-Scout, Gemma-3-12B-it, Gemma-3-27B-it, DeepSeek-R1-14B, DeepSeek-R1-32B, GPT-5-mini), LLMLight~\cite{LLMlight}, Traffic-R1~\cite{zou2025trafficr1reinforcedllmsbring}, and the proposed CuraLight.

\subsubsection{Map traffic flow}Benchmarks are conducted on three real-world networks: \textbf{Jinan} (JN-1/2/3), \textbf{Hangzhou} (HZ-1/2), and \textbf{Yizhuang} (YZ-1/2), with traffic demands (vehicles/half hour) of 2150/2530/2814, 4000/4500, and 8000/10500, respectively.

\subsubsection{Resource Usage}: Dual-socket Intel Xeon Platinum 8457C with a single NVIDIA L20 (48 GiB, CUDA 12.4). Under synchronous batching (avg. batch size $\approx$ 2.85; avg. input/output 960/356 tokens), a total token throughput of 242 tokens/s, and peak GPU memory usage of about 44,200 MiB.

\begin{figure}[h]
\centering
\begin{minipage}[b]{0.32\columnwidth}
  \centering
  \includegraphics[width=\linewidth]{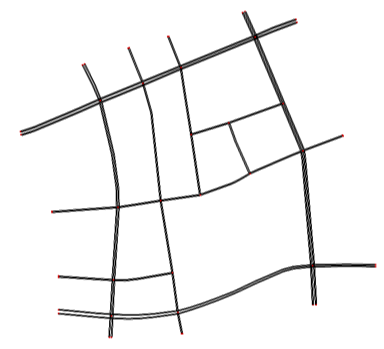}\\[-1mm]
  {\footnotesize (a) Jinan}
\end{minipage}\hfill
\begin{minipage}[b]{0.32\columnwidth}
  \centering
  \includegraphics[width=\linewidth]{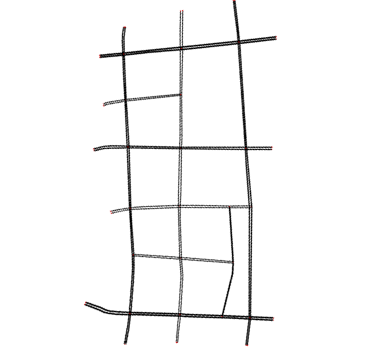}\\[-1mm]
  {\footnotesize (b) Hangzhou}
\end{minipage}\hfill
\begin{minipage}[b]{0.32\columnwidth}
  \centering
  \includegraphics[width=\linewidth]{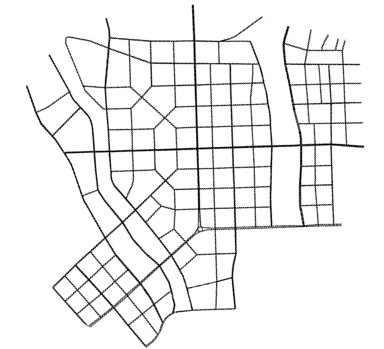}\\[-1mm]
  {\footnotesize (c) Yizhuang, Beijing}
\end{minipage}
\caption{Heterogeneous networks in three real-world scenarios.}
\label{fig:three_scenarios}
\end{figure}

\subsection{Comparison Between CuraLight and Other TSC Methods}
Table~\ref{tab:comparison-all-rounded} shows that CuraLight delivers the most consistent gains across heterogeneous networks and ranks first on most scenario--metric combinations. Compared with FixedTime, MaxPressure, and deep RL baselines such as MPLight and CoLight variants, CuraLight achieves lower travel time, waiting time, delay, and queue length in most settings, indicating stronger adaptability to heterogeneous phase structures and demand patterns.

Compared with the strongest RL baseline UniTSA, CuraLight shows clearer advantages under heavy and heterogeneous traffic. On Hangzhou~2, it reduces ATT from 277.57\,s to 262.14\,s (5.56\%) and AWT from 134.52\,s to 118.80\,s (11.69\%), suggesting a more balanced allocation of service across approaches. An exception appears in Jinan~2: although CuraLight achieves the lowest AQL of 13.06\,m, a 17.03\% reduction from UniTSA (15.74\,m), it does not obtain the best ATT or AWT, indicating a scenario-dependent trade-off between queue dissipation and end-to-end travel time.

\subsection{Generalization Performance in Yizhuang}
\begin{figure}[h]
\centering
\includegraphics[width=\columnwidth]{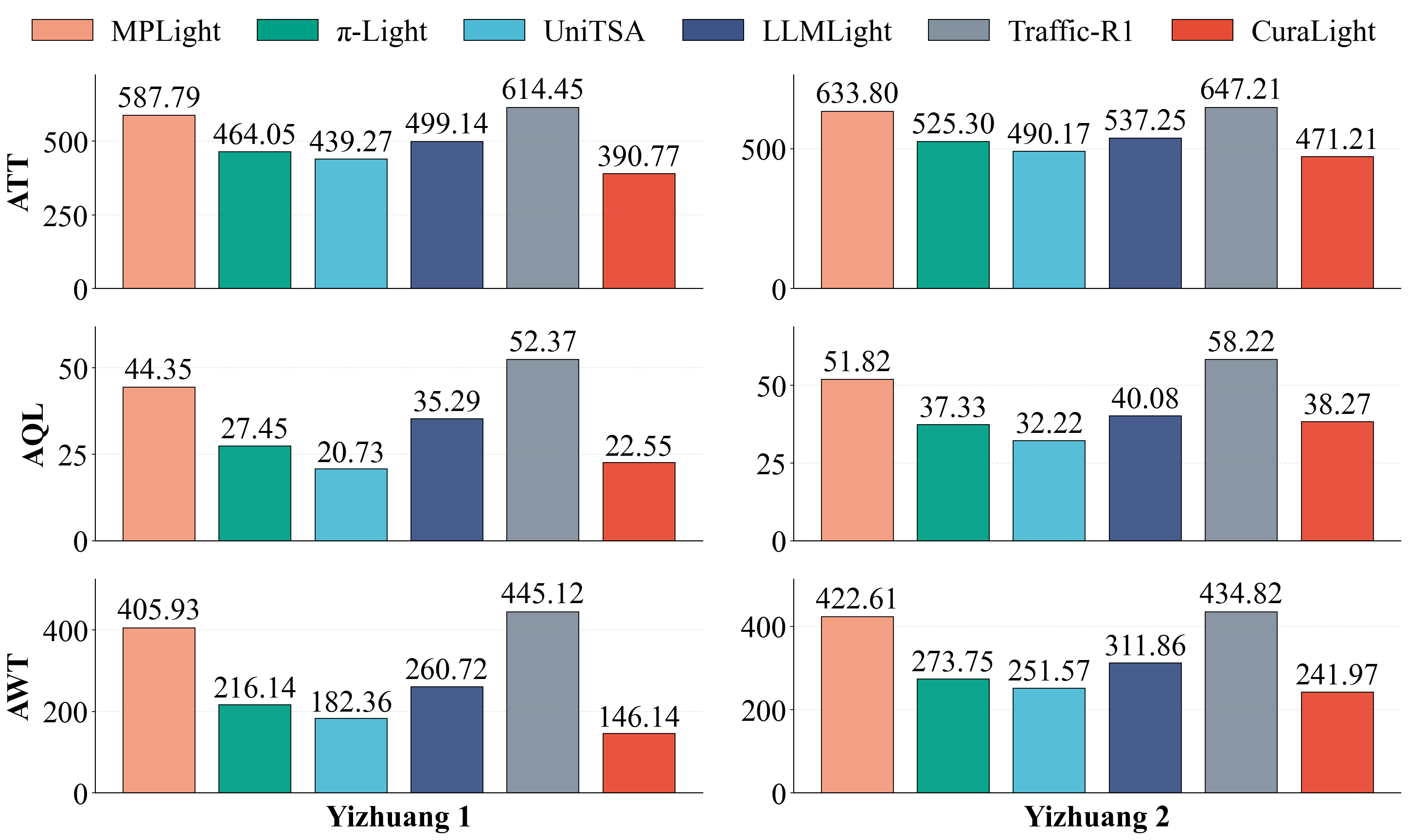}
  \caption{Cross-Network Generalization from Hangzhou 2 to Yizhuang (177 Intersections)}
\label{fig:Ensemble system}
\end{figure}
\textit{Cross-network generalization on Yizhuang.}
Methods are trained on Hangzhou~2 and directly evaluated on the Yizhuang network with 177 intersections under two demand levels (Yizhuang~1/2), without further adaptation. CuraLight shows stronger zero-shot transfer, achieving lower travel time and intersection waiting than UniTSA. For instance, on Yizhuang~1, CuraLight reduces ATT by 11.04\% and AWT by 19.86\% relative to UniTSA. In contrast, AQL is slightly higher, indicating a mild trade-off that prioritizes end-to-end efficiency and waiting reduction over queue storage.

% \subsection{Training CuraLight with RL Filtering and Multi-LLM Ensemble Review}
% \begin{figure}[h]
% \centering
% \includegraphics[width=\columnwidth]{Figures/figure_5.png}
%   \caption{Architecture of the hybrid diffusion + Q-critic agent. 
%   SimpleSSMEncoder encodes the recent state sequence into a context vector $\mathbf{h}_t$; 
%   a conditional diffusion policy generates multiple $(p_k,d_k)$ candidates, 
%   which are evaluated by a Q-critic to select the best action. 
%   Transitions are stored in a sequence replay buffer for off-policy 
%   training of both the diffusion policy and the Q-functions.}
% \label{fig:Ensemble system}
% \end{figure}

\begin{figure}[h]
\centering
\includegraphics[width=\columnwidth]{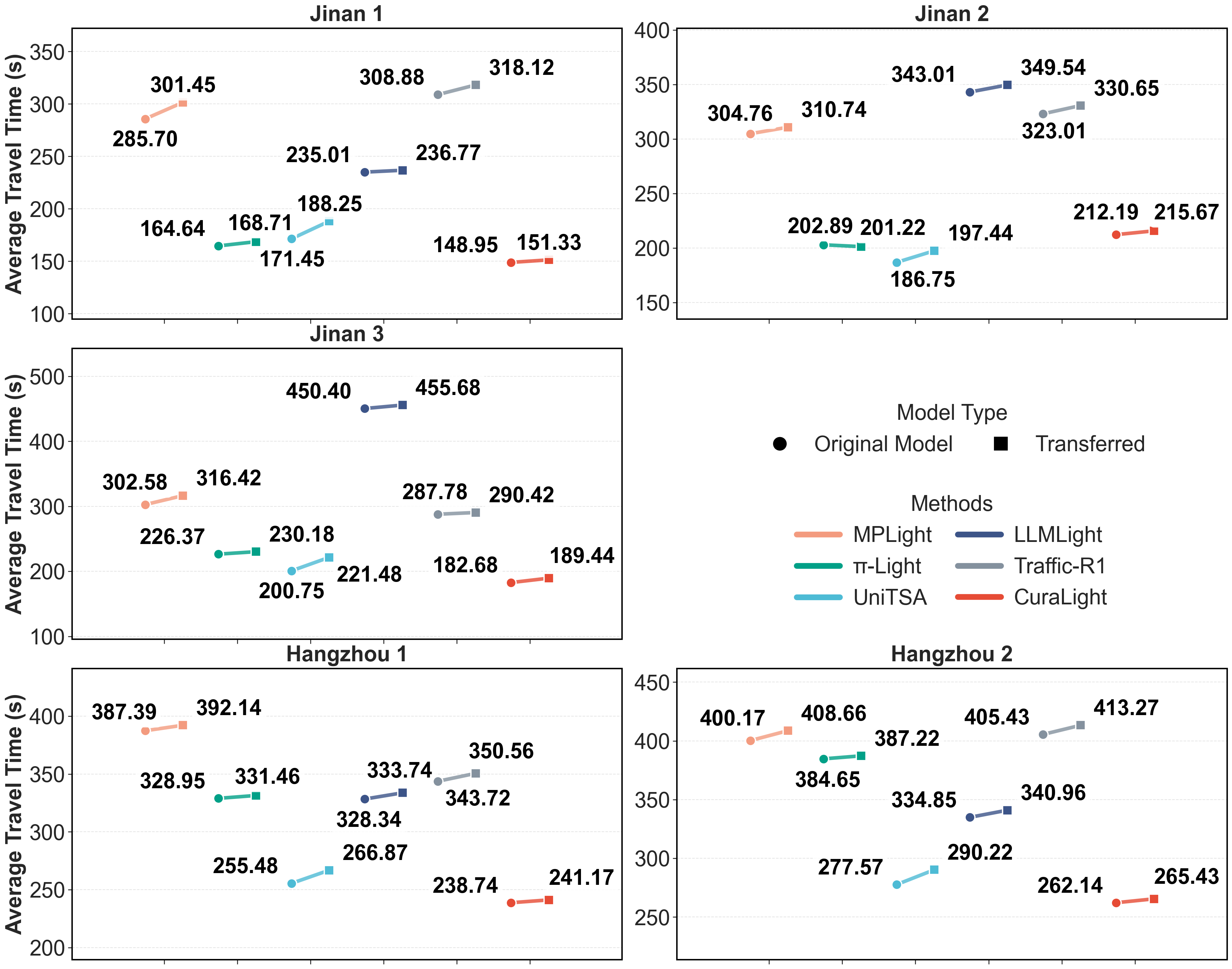}
  \caption{Transferability Comparison on Jinan and Hangzhou}
\label{fig:Ensemble system}
\end{figure}

\subsection{In-scenario transferability}
As shown in Fig.~\ref{fig:Ensemble system}, each \emph{Original Model} (trained on the target scenario) is paired with a \emph{Transferred} model reused from another setting and directly evaluated on the same scenario; the paired markers visualize the ATT change induced by transfer. Overall, reusing a model tends to increase ATT, reflecting the mismatch between the source training distribution and the target traffic dynamics and phase competition patterns. 

CuraLight exhibits noticeably stronger robustness to this shift: its transferred performance degrades only mildly and it remains among the top methods across scenarios. For instance, on Hangzhou~2 the ATT increase of CuraLight is about $1.26\%$, while UniTSA rises by about $4.56\%$, indicating that CuraLight is less sensitive to distribution mismatch. This behavior is consistent with the design that emphasizes RL-filtered candidate timing actions and deliberation-based supervision, which help preserve stable timing preferences when the controller is reused rather than retrained.

\subsection{Ablation Study}

Table~\ref{tab:ablation_single} compares the Gemma-3-12b-it backbone, an imitation-finetuned variant, and the multi-LLM ensemble deliberation system. Imitation fine-tuning reduces ATT, AQL, and AWT relative to the backbone, yielding about 19.3\% average improvement, which suggests that RL-guided interaction trajectories provide effective supervision for learning scenario-specific phase and duration patterns. Adding the multi-LLM ensemble deliberation system brings further gains across all metrics and achieves the best overall results, corresponding to about 28.5\% average improvement over Gemma-3. This indicates that deliberation not only filters clearly inferior timing candidates, but also supplies more structured, consistency-oriented feedback that helps the agent avoid brittle choices under heterogeneous lane configurations. Overall, the ablation supports that imitation establishes a strong base policy, while deliberation-based screening and preference signals offer complementary benefits that further sharpen timing decisions.

\begin{table}[htbp]
    \centering
    \caption{Ablation study on the Hangzhou~2 dataset.}
    \label{tab:ablation_single}
    \resizebox{\columnwidth}{!}{
    \begin{tabular}{lccc}
        \toprule
        \textbf{Model} & \textbf{ATT (s)} & \textbf{AQL (m)} & \textbf{AWT (s)} \\
        \midrule
        Gemma-3-12b-it & 315.42 & 39.78 & 165.25 \\
        Imitation-Finetune & 275.35 & 30.48 & 136.98 \\
        Multi-LLM Ensemble Deliberation System & \textbf{262.14} & \textbf{25.03} & \textbf{118.80} \\
        \bottomrule
    \end{tabular}
    }
\end{table}
\section{Conclusion}
This paper presents CuraLight, an LLM-centered traffic signal control framework that curates high-quality supervision via a diffusion-based RL assistant and a multi-LLM ensemble deliberation system. The RL assistant provides reference timing proposals and preference signals for heterogeneous intersections, while deliberation-based ranking and priority-aware adapter fine-tuning improve the reliability and interpretability of phase and duration decisions. Experiments on real-world networks from Jinan, Hangzhou, and Yizhuang show consistent gains in travel time, queue length, and waiting time over representative transportation, RL, and LLM-based baselines, together with stronger transfer robustness. Future work will focus on reducing deliberation overhead for real-time deployment and extending the framework to broader objectives such as emission-aware control and city-scale coordination.

\bibliographystyle{IEEEtran}
\bibliography{ref}

\end{document}